\documentclass[letterpaper]{article} % DO NOT CHANGE THIS
\usepackage{aaai24}  % DO NOT CHANGE THIS
\usepackage{times}  % DO NOT CHANGE THIS
\usepackage{helvet}  % DO NOT CHANGE THIS
\usepackage{courier}  % DO NOT CHANGE THIS
\usepackage[hyphens]{url}  % DO NOT CHANGE THIS
\usepackage{graphicx} % DO NOT CHANGE THIS
\usepackage{natbib}  % DO NOT CHANGE THIS AND DO NOT ADD ANY OPTIONS TO IT
\usepackage{caption} % DO NOT CHANGE THIS AND DO NOT ADD ANY OPTIONS TO IT
\usepackage{algorithm}
\usepackage{algorithmic}

\usepackage{newfloat}
\usepackage{listings}
\DeclareCaptionStyle{ruled}{labelfont=normalfont,labelsep=colon,strut=off} % DO NOT CHANGE THIS
\floatstyle{ruled}
\newfloat{listing}{tb}{lst}{}
\floatname{listing}{Listing}
\newcommand{\primarykeywords}{
(6) Temporal Planning
}

\usepackage[switch, modulo]{lineno}
\usepackage[utf8]{inputenc}
\usepackage{amsmath}
\usepackage{amsfonts}
\usepackage{graphicx}
\usepackage{color}
\usepackage{booktabs}
\usepackage{xspace}

\newcommand{\fix}[1]{}
\newcommand{\modelname}{CoPEM\xspace}
\newcommand{\sae}{S(AE)$^2$\xspace}
\newcommand{\kdsae}{KDS(AE)$^2$\xspace}
\newcommand{\cope}{CoPE\xspace}
\DeclareMathOperator*{\argmax}{arg\,max}
\DeclareMathOperator*{\argmin}{arg\,min}

\newtheorem{theorem}{Theorem}

\title{Planning and Acting While the Clock Ticks}
\author{
Andrew Coles\textsuperscript{\rm 1},
Erez Karpas\textsuperscript{\rm 2},
Andrey Lavrinenko\textsuperscript{\rm 3},\\
Wheeler Ruml\textsuperscript{\rm 4},
Solomon Eyal Shimony\textsuperscript{\rm 3},
Shahaf Shperberg\textsuperscript{\rm 3}
}
\affiliations {
\textsuperscript{\rm 1}King's College London,
\textsuperscript{\rm 2}Technion,
\textsuperscript{\rm 3}Ben-Gurion University,
\textsuperscript{\rm 4}University of New Hampshire\\
andrew.coles@kcl.ac.uk, karpase@technion.ac.il, andreyl@post.bgu.ac.il, \\
ruml@cs.unh.edu, shimony@cs.bgu.ac.il, shperbsh@bgu.ac.il
}

\begin{document}

\maketitle

\begin{abstract}
Standard temporal planning assumes that planning takes place offline and then execution starts at time 0. Recently, situated temporal planning was introduced, where planning starts at time 0 and execution occurs after planning terminates. Situated temporal planning reflects a more realistic scenario where time passes during planning.
%planning is online.
However, in situated temporal planning a complete plan must be generated before any action is executed. In some problems with time pressure, timing is too tight to complete planning before the first action must be executed. For example, an autonomous car that has a truck backing towards it should probably move out of the way now and plan how to get to its destination later.
In this paper, we propose a new problem setting: concurrent planning and execution, in which actions can be dispatched (executed) before planning terminates.  Unlike previous work on planning and execution, we must handle wall clock deadlines that affect action applicability and goal achievement (as in situated planning) while also supporting dispatching actions before a complete plan has been found.  We extend previous work on metareasoning for situated temporal planning to develop an algorithm for this new setting. Our empirical evaluation shows that when there is strong time pressure, our approach outperforms situated temporal planning.

% Interleaving planning and execution can help solve problems where the some actions must be executed before a complete plan is available, for example, when a truck is backing towards an autonomous car, and it must move out of the way even without a complete plan.
\end{abstract}

\section{Introduction}

Agents operating in the real world, such as robots, must be able to handle the fact that time passes. In temporal planning \cite{DBLP:journals/jair/FoxL03}, the problem formulation accounts for time passing during plan {\em execution}. However, the problem formulation does not account for the time that passes during {\em planning} and thus is suitable for offline planning or for situations where planning time is insignificant compared to execution time.

Situated temporal planning \cite{DBLP:conf/aips/CashmoreCCKMR18} was proposed as a problem formulation in which the passing of time during planning is accounted for. In situated temporal planning, the planner must output a complete plan in a timely manner, that is, before it is too late to execute that plan. This is useful when there are temporal constraints such as deadlines and when planning time might affect the feasibility of meeting such deadlines. For example, situated temporal planning was shown to be useful for online replanning for robots \cite{DBLP:conf/aips/CashmoreCCKMR19}.

However, situated temporal planning is still constrained to output a {\em complete plan} before execution begins. This might be problematic in situations with tight deadlines, which do not allow enough time to find a full plan before the first action is executed. For example, consider a robot assigned to prepare dinner --- a situation marked by a concrete deadline. Despite not having a finalized plan for procuring and assembling all the ingredients, the robot might in some situations be forced to initiate meal preparation, such as boiling water on the stove \cite{pomaine:fct}, to ensure timely completion and meet the dinner deadline.
%consider an autonomous electric vehicle planning an extended journey that involves several stops for charging---a complex planning problem that demands more than just a few seconds to solve. As long as the car can sit idly and plan, there is no need to execute any action. However, there might be a truck backing towards the car, which will collide with the car in 2 seconds. In this case, the correct decision is probably to start driving to get out of the way of the truck, even if the car does not have the full plan for its long trip. \fix{I'm not sure if this example effectively communicates a sense of urgency, as it may lack a clear deadline; the car, for instance, could simply move to a safe parking spot to complete its planning without any imminent time constraints. What about this alternative scenario: consider a robot assigned with a cooking task, specifically preparing a meal for dinner—a situation marked by a concrete deadline. Despite not having a finalized plan for procuring and assembling all the ingredients, the robot must initiate meal preparation, such as boiling water on the stove, to ensure timely completion and meet the dinner deadline.}

%\fix{discuss how this is different from previous `concurrent planning and execution' work.  obsolescence can happen during planning, not only after a plan has been found.  similarly, different from Tianyi's SoCS paper as FACS does not handle action expiration times}

Previous work on combining planning and execution, such as I$_\mathrm{X}$T$_\mathrm{E}$T-$_\mathrm{E}$XEC \cite{DBLP:conf/aaai/LemaiI04} and ROSPlan \cite{DBLP:conf/aips/CashmoreFLMRCPH15} (among others), focused on how to integrate an offline planner with a reactive executive to create an online planning and execution system. This does not address problems with tight deadlines, where plans can become infeasible during the planning process due to a deadline expiring during search.
%In these systems, the decision about whether to execute an action or not is separate from the planning process.
Other work addressed the question of when to commit to dispatching an action during search \cite{DBLP:conf/socs/GuRSSK22}, but this work also does not reason about deadlines and is thus inapplicable to such settings.

In this paper, we formalize the problem of {\em concurrent planning and execution} and describe an algorithm for solving it. Our algorithm builds on the state-of-the-art in situated temporal planning \cite{DBLP:conf/aips/ShperbergCKRS21}, extending it with the option to dispatch an action for execution during search, before a complete plan is found. Although the resulting system contains elements of a traditional executive, we still refer to it as a planner, albeit one that can dispatch actions.

The situated temporal planner we build upon employs a rational metareasoning approach \cite{DBLP:journals/ai/RussellW91} that tries to choose the best computational action based on the information currently available to the planner. Given the inherent complexity of the full metareasoning problem, the planner adopts a pragmatic approach by implementing decision rules derived from a simplified version of the metareasoning problem~\cite{DBLP:conf/aaai/ShperbergCCKRS19}. Prior work has  extended the abstract metareasoning problem to model %handling action costs \cite{DBLP:conf/ijcai/ShperbergCKSR20} and 
concurrent planning and execution (\cope) \cite{DBLP:conf/aaai/ElboherEtAl23}.

Thus, it seems natural to employ the metareasoning approach for concurrent planning and execution in the planner, thereby deriving an algorithm that seamlessly integrates both aspects. Unfortunately, as we discuss later in this paper, 
the \cope~metareasoning model relies on over-optimistic estimates of the probability that some branch of the search tree will ultimately succeed in reaching a solution. Since dispatching an action is irreversible, this can lead to failure. 

Therefore, we present a new simplified metareasoning model, referred to as \modelname~(Concurrent Planning and Execution with Measurements). This model incorporates the understanding that the search process acquires valuable information, enhancing our estimation of the probability of success across various search branches. The new metareasoning model accounts for the value of this information.
% In this paper, we address this problem,
% presenting the \modelname~model that extends COPEC
%by lifting the metareasoning assumption about information gathering, and explicitly assumes that search gathers information according to an (assumed known) sensing model. This has the unfortunate effect of transforming the metareasoning problem from an MDP to a POMDP, making it useless in practice.
While \modelname~yields an intractable POMDP, it clarifies the notion of value of information for additional search effort. This allows us to approximate that value, aiding in the decision-making process of whether to execute actions immediately or await additional information.

Based on these insights and approximations, we present a decision rule for determining whether to search or dispatch an action. This decision rule is integrated into the situated temporal planner of \citet{DBLP:conf/aips/ShperbergCKRS21}, yielding a dispatching planner for concurrent planning and execution. An empirical evaluation shows that this system outperforms situated temporal planning whenever time pressure is tight.

%\fix{unrelated work: planning and execution systems (IxTeT)}

% We then describe how to modify an existing situated temporal planner \cite{DBLP:conf/aips/ShperbergCKRS21} to account for the effects of executing an action on the search. 

% We also describe a metareasoning scheme, based upon metareasoning for situated temporal planning \cite{DBLP:conf/aaai/ShperbergCCKRS19,DBLP:conf/aips/ShperbergCKRS21} for deciding {\em when} to execute an action before we have a complete plan, and {\em which} action to execute.

\section{Problem Statement}
%: Concurrent Planning and Execution}
\label{s:problem_statement}

We define {\em concurrent planning and execution} in a manner similar to situated temporal planning \cite{DBLP:conf/aips/CashmoreCCKMR18}: as propositional temporal planning with Timed Initial Literals (TILs), formalizable in PDDL 2.2 \cite{cresswell:ptl,edelkamp2004pddl2}. The sole distinction between concurrent planning and execution and situated temporal planning is the ability to execute an action before we have a complete plan, which is formalized by  slightly different constraints on {\em when} the output is produced.

Formally, a concurrent planning and execution problem $\Pi$ is specified by a tuple $\Pi = \langle F, A, I, T, G \rangle$, where
% \begin{itemize}
% \item 
$F$ is a set of Boolean propositions that describe the state of the world,
% \item 
$A$ is a set of durative actions, with $a \in A$ composed of
  %\begin{itemize}
  % \item Minimum duration $\mathit{dur}_{\min}(a)$ and maximum duration $\mathit{dur}_{\max}(a)$, both in $\mathbb{R}^{0+}$ with $\mathit{dur}_{\min}(a) \leq \mathit{dur}_{\max}(a)$,
    %\item 
    a duration, $\mathit{dur}(a) \in \mathbb{R}^{0+}$,
  start condition $\mathit{cond}_{\vdash}(a)$, invariant condition $cond_{\leftrightarrow}(a)$, and end condition $cond_{\dashv}(a)$, all of which 
  are subsets of $F$. The effects are given by 
  % \item 
  start effect $\mathit{eff}_{\vdash}(a)$ and end effect $\mathit{eff}_{\dashv}(a)$, both of which specify which propositions in $F$ become true (add effects), and which become false (delete effects).
  %\end{itemize}
% \item 
$I \subseteq F$ is the initial state, specifying exactly which propositions are true at time 0. 
% \item 
$T$ is a set of timed initial literals (TIL). Each TIL $l = \langle \mathit{time}(l), \mathit{lit}(l) \rangle \in T$ consists of a time $\mathit{time}(l)$ and a literal $\mathit{lit}(l)$, which specifies which proposition in $F$ becomes true (or false) at time $\mathit{time}(l)$, and
% \item 
$G \subseteq F$ specifies the goal, that is, propositions we want to be true at the end of plan execution.
% \end{itemize}

As in situated temporal planning, TILs are seen as temporal constraints in {\em absolute time since planning started}.
However, unlike situated temporal planning, where we require generation of a full plan before execution begins, in concurrent planning and execution we allow our algorithm to dispatch an action, even before a complete plan is available.

Formally, our algorithm outputs a sequence $\sigma$ of pairs $\langle a, t_a \rangle$, where $a \in A$ is an action and $t_a \in \mathbb{R}^{0+}$ is the time when action $a$ is to start. The first requirement is that this sequence of actions forms a valid solution for the planning problem $\Pi$, that is, that all conditions hold at their respective time points, and that the plan achieves the goal, exactly as in standard temporal planning.  In more detail, we define a valid schedule by viewing it as a set of instantaneous {\em happenings} \cite{DBLP:journals/jair/FoxL03} that occur when an action starts, when an action ends, and when a timed initial literal is triggered. For each pair $\langle a, t_a \rangle$ in $\sigma$, we have action $a$ starting at time $t_a$ (requiring $\mathit{cond}_{\vdash}(a)$ to hold a small amount of time $\epsilon$ before time $t_a$, and applying the effects $\mathit{eff}_{\vdash}(a)$ right at $t_a$), and ending at time $t_a+\mathit{dur}(a)$ (requiring $cond_{\dashv}(a)$ to hold $\epsilon$ before $t_a+\mathit{dur}(a)$, and applying the effects $\mathit{eff}_{\dashv}(a)$ at time $t_a+\mathit{dur}(a)$). 
For a TIL $l$ we have the effect specified by $\mathit{lit}(l)$ triggered at $\mathit{time}(l)$.  Finally, we require the invariant condition $\mathit{cond}_{\leftrightarrow}(a)$ to hold over the open interval between $t_a$ and $t_a+\mathit{dur}(a)$, and the goal $G$ to hold after all happenings have occurred. 

However, uniquely to our problem, we must also ensure our algorithm does not dispatch actions in the past. Thus, we annotate the output from our algorithm with the time each pair was output. That is, we treat the output as a sequence $(t_1, \langle a_1, t_{a_1} \rangle), \ldots,  (t_n, \langle a_n, t_{a_n} \rangle)$, where $t_1 \leq t_2 \leq \dots t_n$, indicating that the pair $\langle a_i, t_{a_i} \rangle$ was output at time $t_i$. The requirement is that $t_i \leq t_{a_i}$---meaning our algorithm commits solely to dispatching actions now or in the future. We remark that there is never any theoretical benefit in committing to an action in the future, and thus typically we would expect to see $t_i=t_{a_i}$. However, practical considerations might interfere with this, as we discuss below.
For comparison, in situated temporal planning, the requirement is $\forall i: t_n \leq \min t_{a_i}$. This formulation generalizes cases where the planner outputs a complete plan at once, in which case $t_1 = \dots = t_n$.
If our algorithm is able to emit a sequence of actions that forms a valid plan, without violating the laws of space-time by dispatching actions in the past, we call it {\em successful}.

Consider the aforementioned example of a robot preparing dinner. In this scenario, a TIL $l$ is utilized to enforce that the meal must be ready before dinner time, $time(l)$. If the planning time is substantial, $t_n$ will approach $time(l)$. Without concurrent dispatching, all actions must be executed within the timeframe $t_n - time(l)$, which could be infeasible, given the time-consuming nature of cooking. However, in our concurrent setting, the sole requirement is for the final action to be executed after $t_n$, significantly enhancing the feasibility of the task.
For example, it is beneficial to dispatch long actions (such as boiling a large pot of water) early, as these will no longer be constrained to fit within the time window between when planning ends and the deadline.

%\section{Background: Abstract Models for Metareasoning in Situated Temporal Planning}
\section{Prior Work}%: Situated Temporal Planning, Metareasoning}

%\fix{say something about forward search somewhere}

%Having described the problem we are addressing, we now describe the previous work we build upon. As this work spans multiple papers, we present only the most relevant parts in detail.
The situated temporal planner we build upon \cite{DBLP:conf/aips/ShperbergCKRS21} uses the OPTIC planner \cite{DBLP:conf/aips/BentonCC12}. OPTIC applies heuristic forward search in the space of sequences of happenings (snap actions). Some temporal planners use different types of search techniques, such as constraint propagation \cite{DBLP:journals/ai/VidalG06} or compilation to SAT \cite{DBLP:journals/jair/RankoohG15}. These planners were designed for offline planning. However, when we want to consider dispatching an action, it is much easier to do so in the context of a current state of the world, and thus relying on forward search seems to be the most natural approach.
Of course, it is possible to adapt other forward search planners \cite{DBLP:journals/jair/GereviniSS03,DBLP:conf/aips/Vidal04,DBLP:conf/aips/EyerichMR09} with the ability to dispatch actions. However, the planner we build upon has most of the machinery needed for metareasoning, which we describe next, thus making our job easier.

%\fix{say something about other forward search planners (TFD, YAHSP)}

\subsection{Abstract Metareasoning}

Rational metareasoning \cite{DBLP:journals/ai/RussellW91} provides a way to choose among different computational actions. The decision problem that metareasoning addresses is called the meta-level decision problem, and it can be formalized as an MDP, or as a POMDP when we have partial information.
%Rational metareasoning  addresses a decision problem, called the meta-level decision problem, which attempts to choose which computational actions to execute. 
Computational actions in our setting include expanding a search node or dispatching an action, making the meta-level decision problem rather complicated.% and entails how to prioritize search and whether to dispatch actions before search terminates.

\citet{DBLP:conf/aaai/ShperbergCCKRS19} address part of this metareasoning problem (excepting action dispatching), by abstracting from the intricacies of the plan state representation and search process.   They model the problem using $n$ processes, denoted $p_1, \ldots, p_n$, where each process is dedicated to searching for a plan (each process can be thought of as representing a search node on the open list). Each process is described by a probabilistic performance profile, modeled by a random variable (RV) indicating the probability of process $p_i$ terminating successfully given processing time $t$; $M_i$  denotes the Cumulative Distribution Function (CDF) of this RV.

Processing must terminate before a deadline, which may be unknown during planning, and is thus also assumed to be a random variable. The CDF of the deadline by which time process $p_i$ must terminate in order for its solution to be usable is denoted by $D_i$. Note that the deadline is with respect to `wall clock' time (total time allocated to all processes), while $M_i$ counts `CPU time' (time allocated to $p_i$).

Under the assumption that information about the true deadline and processing time of process $p_i$ is only available when that process terminates, the problem is to find an optimal policy for scheduling processing time for all the processes, so that the probability that some process $p_i$ will deliver a plan before its deadline is maximal. A slightly simplified version of this problem, when time is discrete (assumed to be integer-valued) is known as Simplified Allocating planning Effort when Actions Expire, or \sae~for short. Solving \sae~optimally was shown to be NP-hard, but if the deadlines are known (called \kdsae~, with KD standing for Known Deadlines), the problem can be solved optimally in pseudo-polynomial time by dynamic programming \cite{DBLP:conf/aaai/ShperbergCCKRS19,DBLP:conf/aips/ShperbergCKRS21}.

As even solving the simplified problem using the pseudo-polynomial algorithm is too expensive, previous work relies on a simplified decision scheme called Delay Damage Aware (DDA), which is based on ideas used in the optimal \kdsae.
The DDA scheme relies on the log-probability of failure (LPF) of allocating $t$ consecutive units of computation time to process $i$, starting at time $t_b$, denoted $LPF_i(t, t_b)$. To compute the LPF, we first compute 
the probability that process $i$ finds a timely plan when allocated $t$ consecutive time units
beginning at time $t_b$, which is:
\begin{equation}
  s_i(t,t_b) = \sum_{t'=0}^t m_i(t')  \cdot (1-D_i(t'+t_b))
  \label{eq:LPFs}
\end{equation}
where $m_i(t)=M_i(t) - M_i(t-1)$, i.e. the PMF of $M_i$.
The choice to use the log of the probability of failure allows us to treat it like an additive utility function, thus we define $LPF_i(t, t_b) = log(1-s_i(t, t_b))$.

The DDA scheme allocates chunks of $t_u$ computational time units (where $t_u$ is a hyperparameter).  The utility of a process $i$ is defined by the log-probability of failure of allocating computation time to process $i$ in the next chunk (starting at time $t_u$, with a discount factor of $\gamma$) minus the log-probability of failure of allocating time to process $i$ now, thus accounting for the urgency of the process. The amount of computation time to use in the utility calculation is chosen by the {\em most effective computation time} for process $i$ starting at time $t_b$, defined as $e_i(t_b) = \arg\!\min_t \frac{LPF_i(t,t_b)}{t}$, that is, the time allocation is chosen by its marginal gain. Putting this all together, the DDA scheme allocates the next unit of computation time to the process $i$ with maximal 
\begin{equation}
Q(i) = \frac{\gamma \cdot LPF_i(e_i(t_u), t_u)}{e_i(t_u)}- 
\frac{LPF_i(e_i(0), 0)}{e_i(0)}
\end{equation}

\subsection{MR in Concurrent Planning \& Execution}

An abstract metareasoning model for Concurrent Planning and Execution, called \cope, was presented by
%Subsequent metareasoning models also support concurrent planning and execution. The COncurrent Planning and Execution (\cope) 
\citet{DBLP:conf/aaai/ElboherEtAl23}. The \cope~metareasoning model extends \sae~by assuming that each process $p_i$ has already computed a plan prefix $H_i$ consisting of some actions (for example, if $p_i$ represents a node on the open list, $H_i$ are the actions leading from the initial state to that node). One is allowed to start executing actions from some $H_i$ before planning terminates (and concurrently with planning), but doing so invalidates all processes that have initial actions inconsistent with the prefix of $H_i$ already executed. The requirement now is to have at least one still valid process $p_i$ complete its computation {\em and} execute its $H_i$ before some {\em induced deadline} $ID_i$ (the induced deadline can be computed from $D_i$ and the action durations, but we omit the details here for the sake of brevity). Note that a \cope\ problem instance where all $H_i$ are empty is also an \sae~instance.

For the special case of \cope\ where the induced deadlines are known (denoted KID\cope), it is possible to reduce the problem to multiple instances of $KDS(AE)^2$,
which in turn can be solved in pseudo-polynomial time by dynamic programming \cite{DBLP:conf/aaai/ElboherEtAl23}. This is done by choosing an execution time for
%fixing the start of the 
all the actions in some $H_i$. The function that defines the execution time for each action in $H_i$ is called an {\em initiation function}, and we denote it by $f$.
%, i.e. providing an {\em initiation function} $f$ from actions in $H_i$ to time. 
The semantics is that each action $a\in H_i$ is executed beginning at $f(a)$, unless $a$ becomes redundant because
a complete timely plan that does not use $a$ is found before $f(a)$.
Recall that when 
%Each such 
action $a$ is executed, it invalidates any process $j$ that has an $H_j$ inconsistent with $a$.
Thus, given $f$, one can define an {\em effective deadline} $d_i^{\mathit{eff}}$ for each process $i$
as the minimum of $ID_i$ and the execution time $f(a)$ for the first action inconsistent with $H_i$. Using the effective deadlines, we get, rephrasing Theorem 6 from \cite{DBLP:conf/aaai/ElboherEtAl23}:

\begin{theorem}
Given a \cope~ problem instance $I$
and an %execution 
initiation function $f$,
using the effective deadlines
as computation deadlines (and subsequently ignoring the $H_i$) defines a $KDS(AE)^2$ instance $KDS(I,f)$.
The optimal policy for $I$
restricted to action execution according to $f$ is equal to the optimal computation policy for
$KDS(I,f)$.
\end{theorem}

Solving the resulting $KDS(AE)^2$ instances for  all possible initiation functions $f$ and picking the one with the highest success probability
is an algorithm for optimal solution of the KID\cope ~ instance.
Obviously, the complexity of the above algorithm is exponential in  $\max_i |H_i|$, but special cases exist where the number of possible $f$ is polynomial
\cite{DBLP:conf/aaai/ElboherEtAl23}. We use a similar technique below.

\subsection{Metareasoning in a Planner}
\label{sec:metareasoninginanplanner}

So far, we have discussed abstract metareasoning models. Integrating these into a planner is not trivial. We now explain briefly how the situated temporal planner we build upon \cite{DBLP:conf/aips/ShperbergCKRS21} uses the DDA decision rule in practice.

First, there are several adaptations to the node expansion process itself, accounting for TILs that occur during planning and pruning nodes for which it is already too late to start executing \cite{DBLP:conf/aips/CashmoreCCKMR18}. Second, to use the DDA decision rule, the planner must estimate $M_i$ and $D_i$, which are used to compute the LPF.

An admissible deadline for node $i$ can be found by building a Simple Temporal Problem (STP) for the plan prefix $H_i$, solving it to find the latest feasible timestamp $tmax$ of each step, and taking the minimum of these across all steps:
$$\mathit{latest\_start(H_i)} = \mathop{\min}_{a_j \in H_i} \mathit{tmax}(a_j)$$
In practice, a more informative but inadmissible estimate is found based on the temporal relaxed planning graph (TRPG) \cite{DBLP:conf/aips/ColesCFL10} heuristic by additionally including the relaxed plan $\pi_i$ in this STP. This estimate is called the {\em estimated latest start time} for node $i$ and is used as a known deadline (that is, $D_i$ assigns a probability of 1 to this being the deadline).

To estimate $M_i$, the planner relies on the notion of expansion delay \cite{dionne:das}---the average number of expansions between when a node is generated and when it is expanded. A distribution is built around this estimate based on statistics collected during search; the details of how this is done are omitted for the sake of brevity.

With these estimates of $M_i$ and $D_i$, the planner can compute $Q_i$ for each node on the open list and sort the open list based on $Q$. 
Since the DDA scheme is based on allocating $t_u$ units of computation time to the chosen process $i$, the planner performs $t_u$ expansions in the {\em subtree} rooted at $i$; after $t_u$ expansions, the non-expanded (frontier) nodes in this subtree are added to the open list and another state is chosen according to $Q$. Additionally, as the estimates for $M_i$ could change (because the statistics collected during search to estimate $M_i$ change), the $Q$ value is recomputed for all the nodes on the open list every $t_u$ expansions.

So far we discussed how DDA, a metareasoning scheme for \sae, was integrated into a planner. 
The issue with integrating \cope~into a planner is that the planner's estimates of the $M_i$ and $D_i$ distributions can be quite far off the mark, especially early in the search;  thus might lead to wrong decisions. In situated temporal planning, this is not critical, because a wrong decision wastes only some search effort (a few node expansions). 
% and which in many cases can be overcome later on.
% However,
% the decisions in the implementation of the planner based on \sae~a greedy rule is used to decide on what branch of the search tree to expand for a while. This update and 
% violation is not critical, 
However, in concurrent planning and execution, a wrong decision to dispatch an action can frequently be fatal. %Yet when the planner estimates $M_i$ and $D_i$ based on scant evidence, the wrong decision can easily be made.
Therefore, our first step is developing an improved metareasoning model, which accounts for the  information gathered by `measurements' during search.

% \fix{explain more about how previous work estimates $M_i$, $D_i$, LPF, search and subtree focus, compatible/incompatible $H_i$}

\section{Metareasoning with Measurement Model}

% Making such a decision based on scant evidence seems to be what made applying \cope~fail. Indeed, refusing to dispatch an action in a poorly developed subtree, and giving priority to searching a subtree corresponding to an action we are about to dispatch
% if the subtree is small, immediately improves the results, However, this is an ad-hoc solution not well justified by theory. 

We now present our new \modelname~abstract metareasoning model,  extending \cope~by explicitly making the more realistic assumption that computation actions deliver information that can update the distributions. %rather than only upon termination of a process.
\modelname is obviously at least as computationally hard to solve optimally as \cope, 
%the PSPACE-hard ACE, 
so except for restricted cases, we do not attempt to solve it optimally. Its main role is to specify a notion of what it means to be optimal in concurrent planning and execution. Nevertheless, we leverage ideas from optimal solutions of the restricted cases towards an actual implementation solving the concurrent planning and execution problem from Section \ref{s:problem_statement}.

% if the subtree is small, immediately improves the results, However, this is an ad-hoc solution not well justified by theory. 

%making a decision to dispatch an action based on scant evidence is g

\subsection{The \modelname~Model}

As in the \cope~model, we have $n$ processes $p_1...p_n$, all searching for a plan starting at a known initial state.
Each process $p_i$ has already computed an initial action sequence $H_i$, where each action in $H_i$ has a specific time window for execution.
For each process we have a random variable $M_i$, a performance profile determining how long it needs to compute until termination. Random variable $D_i$ is a distribution over the induced deadline of process $p_i$: the time by which the last action in $H_i$ and the rest of any solution found by process $p_i$ must be executed in order to be successful.
%Additionally, we have $C_i$, random variables, each  determining the eventual plan cost of plan $i$. 
In general, the random variables may be dependent.

There are three types of actions: real-world actions to be dispatched corresponding to the next action from some $H_i$, a computation action $c_i$ allocating a processing time unit to process $p_i$,
or commit to
a complete correct plan
found by any terminated process $p_i$. 
All actions are non-preemptible and mutually exclusive, except that computation actions can be run concurrently with at most one real-world action.
A computation action $c_i$ can make the process terminate (with probability determined by $M_i$), in which case the process delivers a solution and its true deadline is revealed.

Up to now, this is the same as the \cope~model. 
%($n$ processes, with completion time distribution, deadline distribution, and cost distribution). 
However, in \modelname~the distributions over the random variables $M_i$ and $D_i$ are just priors: a computation $c_i$ also delivers  an observation $o\in O_i$ (for some observation space $O_i$) as evidence that affects the posterior distributions over the random variables according to a known {\em measurement model}. In other words, the action $c_i$ has a stochastic effect on what the $M_i$ and $D_i$ distributions would be in the next (belief) state, after updating them according to the observation $o$.

To parameterize a restriction on the model complexity, we define a parameter $K$, the time at which we no longer allow a real-world action to be dispatched `early' (i.e. before planning terminates), and a parameter $L$, the last time at which observations can be received (when a computation action $c_i$ performed at time $t$ causes the observation $o$ to be received in time to make the decision at time $t+1$). The \cope~model is a special case of \modelname~where $L=0$ and $K=\infty$.

%The above distributions and sensor models should be learned from observation of search processes. This learning issue is beyond the scope of this paper.
%and is explored separately \cite{ }. 
We assume here that the measurement model is known and that we can perform Bayesian updating on the runtime and deadline distributions. Under this assumption, like many metareasoning problems, the \modelname~model is a POMDP, the solution of which is intractable: potentially exponential in the number of time units in the model, thus certainly not optimally solvable in real-time. We therefore examine some special cases, and leverage their solutions towards a greedy decision-making algorithm to handle \modelname~in practice.

\subsection{Basic Tractable Case}

% \fix{Write formal definition of POMDP in appendix, include brief intuition here}

We begin with the restricted \modelname case that we call {\em fully resolved}.
Given a \modelname instance $I$ 
and a commitment to execute all the actions in some $H_i$ at certain times (an initiation function $f$),
$(I, f)$ is fully resolved if
the induced deadlines are known and no further information can be obtained by
computations about any of the $M_i$ distributions.

\begin{theorem}
If $(I, f)$ is fully resolved, the optimal policy for $I$
under initiation function $f$
is equal
to that of an equivalent
$KDS(AE)^2$ instance $I(I,S)$.
\end{theorem}

{\bf Proof}:
If no further information on the $M_i$ can be obtained by computation, and the induced deadlines are known, we have an instance of KID\cope . Since we also have a fixed initiation function $f$, by Theorem 1 we can construct an instance $KDS(I,f)$ whose optimal computation policy
is also optimal for $I$.
$\Box $

%Despite the fact that a fully resolved instance
%requires multiple decisions, it enables a pseudo-polynomial optimal solution algorithm
%rather than an exponential one, as $KDS(AE)^2$ can be addressed using dynamic programming within pseudo-polynomial time.
Leveraging this tractability property to solve instances of \modelname would be advantageous. However, its application is nontrivial, since a \modelname instance is not fully resolved, even for $K=L=0$ and with known induced deadlines, the first action in any of the $H_i$ could be dispatched at $t=0$ (or none), i.e. we do not have an initiation function.

Nevertheless, consider a \modelname
instance restricted to $K=L=0$,
known induced deadlines,
and $|H_i|=1$ for all $i$.
Opting not to dispatch an action at $t=0$,
we are not allowed to dispatch any actions
early subsequently.
Since no further information on $M_i$ can be acquired, effectively we have a fully resolved equivalent instance.
Consequently, we can formulate an equivalent \kdsae~ instance denoted as ${\cal I}_0$.
Likewise, when opting to dispatch the only available action $a\in H_i$ for some $1 \leq i \leq n$, $K=0$ implies $f(a)=0$, so again we have a fully resolved instance,
We denote
the respective
\kdsae instance by ${\cal I}_i$.

Once we generate these $n+1$ \kdsae~instances, we solve each of them optimally in pseudo-polynomial time by dynamic programming, obtaining a value $v_i$ (probability of success) for each of them. Our dispatch decision is then given by $\argmax_{i=0}^n v_i$, where a decision of $i>0$ indicates dispatching the first action in $H_i$, and a decision of $i=0$ indicates not dispatching any action.
This scheme, henceforth called Solve00$(\cdot)$, returns the best $v_i$
and (optionally) the solution for the respective \kdsae~instance.

\begin{theorem}
Given an instance $I$ of
\modelname with $K=L=0$, known induced deadlines,
and $|H_i| \leq 1$ for all $1 \leq i \leq n$, Solve00$(I)$
yields the optimal policy for $I$.
\end{theorem}
%\fix{prove formally this gives an optimal solution}.
Note that the above method is applicable for $|H_i| > 1$, but the DP solution to the \kdsae~instance is not guaranteed to be an optimal \modelname solution beyond $|H_i|=1$.

%Thus we get the following solution algorithm for \modelname~$K=1,L=0$ instances:

% \begin{enumerate}
%      \item Set $i_{opt}=0$, 
%      and let ${\cal I}_0$ be the \kdsae~equivalent instance for dispatching no action.
%      \item Solve ${\cal I}_0$ using DP, find its success probability $P_0$,
%      and set $P_{opt}=P_0$
%     \item For $i=1$ to $n$
%     \begin{enumerate}
%         \item Create \kdsae~instance ${\cal I}_i$ equivalent for dispatching action in $H_i$.
%         \item Solve ${\cal I}_i$, and if its success probability $P_i>P_{opt}$, update: $P_{opt}=P_i, i_{opt}=i$.
%     \end{enumerate}
%     \item If $i_{opt}>0$, dispatch action from $H_i$.
%     \item
%     Allocate computations according to the optimal policy found for \kdsae~instance, ${\cal I}_{i_{opt}}$.
%  \end{enumerate}

%\subsection{Details of reduction to \kdsae~}

\subsection{Extension to Incorporate Measurements}

The above restrictions can be relaxed to scenarios with a bounded number $L$ of informative (w.r.t. $M_i$) computation actions and $K$ dispatch decisions, as long as our observation space $O_i$ for each computation action $c_i$ is finite.  
Let $O_{max}=\max_i |O_i|$.
We focus on $K=L=1$, which is similar in spirit to the Russell and Wefald myopic assumption \cite{DBLP:journals/ai/RussellW91,DBLP:journals/tsmc/TolpinS12}.
%That suffices for gaining valuable insights into combining metareasoning with heuristic search for real concurrent planning and execution problems.%in a planner.

%This requires a measurement model for $M_i$.%, which is not easy to obtain. 
%In addition, the myopic assumption that only one measurement will be made is risky, and although it can be overcome by the ``blinkered'' scheme \cite{DBLP:journals/tsmc/TolpinS12} which considers value of multiple measurements, this scheme requires a well-understood measurement model which we do not have at present.

%\fix{say somewhere that updating with observation $O$ gives us a distribution $M_i$ which gives the exact value for runtime}
%Instead, we suggest to compute the value of perfect information (VPI), that is, under the metareasoning 

%This is an overestimate, so if the VPI of all possible computations is lower than its cost (the difference is called the {\em net VPI}), then we can safely choose to dispatch an action. Note that the cost of a computation action could be negative, as it costs time but also has a probability of terminating successfully.

We also assume here known induced deadlines, and jointly independent $M_i$ distributions.
For clarity, we further assume
that a computation action $c_i$ yields perfect information, i.e. an
observation $o$ equal exactly to the remaining computation time for process $p_i$. Note that this assumption can be easily adjusted should an alternative measurement model for each computation action be available.
With these assumptions, the only uncertainty is in the $M_i$, and in observations to be received due to the first computation action. 
%The perfect information assumption entails that computation $c_i$ delivers an observation $o$ that reveals the true runtime of process $p_i$.
%Let $m_i(t) := M_i(t) - M_i(t-1)$ be the Probability Density Function (PDF) of $M_i$.
Perfect information implies that computation $c_i$ will observe $o=t$ with probability $P(o=t) = m_i(t)$. 

% (Extending to imperfect observation models.) Note that we could use any 
% measurement model
% $P(O|m_i)$ (where $m_i$ is the unknown true value), if available,
% In this case we have 
% $P(O=t)=\sum_{t'}m_i(t')P(O=t|m_i=t')$ which reduces to the above simpler equation $P(O=t) = m_i(t)$ with perfect observations.

For conciseness and concreteness, we define notation for \modelname~and \kdsae~instances created due to dispatching decisions and/or observations, and how they are defined.
Let ${\cal I}$ be the original \modelname~instance with $K=1, L=1$. 
We break this down into two subcases: 

\noindent {\bf Not dispatching at time 0:}
Executing a computation $c_i$, we then get an observation $o\in O_i$, and in each such case we need to optimize a new \modelname~instance, where the current time is $t=1$. 
We can shift the origin of the $t$ axis to 1 and treat the resulting state as a $K=L=0$ instance.
Let us denote each such problem instance by  ${\cal I}(c_i, o)$. 
Denote by subscripting ${\cal I}_i$ the problem instance resulting from dispatching decision $i$, \fix{confusing since this is the non-dispatching case?}
which is dispatching the (first and only) action in $H_i$, or deciding not to dispatch at this time for $i=0$.
An instance ${\cal I}(c_j, o)$ after a dispatch decision, denoted
${\cal I}_i(c_j, o)$
is now a \kdsae~instance, since no
more observations can be received, and no additional early dispatch decisions are allowed.
Note that the distribution model of
${\cal I}_i(c_j, o)$ is the same as that of
${\cal I}_i$ for all processes except $m_j$,
which is replaced by the 
%cdf of the 
degenerate $\delta$-distribution $\delta(o)$.% (where $\delta$ is the delta function).

\noindent {\bf Dispatching at time 0:}
Alternatively, we can decide on an early dispatch
in ${\cal I}$ at time $t=0$, resulting in
an instance ${\cal I}_i$.
Note that the result is not a \kdsae~instance. However, since
no further actions can be dispatched, each ${\cal I}_i$
is essentially a $K=0, L=1$ instance.
But after the next computation $c_j$ is done, we receive the observation $o$
and get a \kdsae~instance again: not
the same as ${\cal I}_i(c_j, o)$ because here
an action from $H_i$ has been dispatched at time 0, rather than 1, so we denote this by
${\cal I}^0_i(c_j, o)$.
To find the probability of success of the optimal policy, compute:
\begin{equation}
Solve01({\mathcal I}_i) = \max_j \!\!\!\!\!\!\! \sum_{o\in support(m_j)}
\!\!\!\!\!\!\!\!\!\! m_j(o) Solve00(\mathcal{I}^0_i(c_j, o))
\end{equation}
and any $c_j$ achieving the maximum is optimal.% The full pseudo-code for Solve01 can be found in the appendix.

All in all, the $K=L=1$ case is handled by pseudo-polynomial time
computation of the following equations, denoted as Solve11.
For all $1\leq i\leq n$:
\begin{equation}
P_i = \sum_{o\in support(m_i)}
m_i(o) Solve00(\mathcal{I}_i(c_i, o))
\end{equation}
\begin{equation}
P^\prime_i = Solve01(\mathcal{I}_i)
\end{equation}
Then, select the policy corresponding to the maximum value of all $P_i$ (compute at time 0 policies)
and all $P^\prime_i$
(dispatch at time 0 policies).
\fix{(Pseudo-code is provided in the supplement.)}
Essentially we evaluate a depth 2 expectimax tree,
with leaf nodes being \kdsae ~ instances, and return the best policy and its corresponding probability of success.
The overall complexity of this scheme is $O(T^2n^4O_{max})$
where $T$ is the number of time steps. $T^2n^2$
is the time taken to solve a single \kdsae instance by dynamic programming, and $n^2O_{max}$ simply counts the loops to compute the equations. %The complete pseudo-code for Solve11 is available in the appendix. 

In principle, this scheme can be extended to greater $K$ and $L$,
paying a factor $n^{L+K}O_{max}^L$ by optimizing over all possible action and observation sequences of length $L$ and dispatching decision sequences of length $K$. However, since even the $K=L=1$ case, despite being tractable, is too heavy to use in metareasoning, this is not examined here.

\subsection{Leveraging the Tractable Case in Practice}

%We also develop a greedy decision rule, which captures the intuition that refusing to dispatch an action in a poorly developed subtree (meaning the distribution estimates there are likely to be wrong) is dangerous. Instead, we give priority to searching a subtree corresponding to an action which seems like it should be dispatched, but this decision is based on little evidence. Using our model, we show that a variant of this greedy scheme is correct from a theoretical perspective as well.

In applying the \modelname~model to concurrent planning and execution, obviously our metareasoning assumptions of independence, known induced deadlines, and the myopic $K=1, L=1$ assumptions do not hold.
Nevertheless we can use this as a first approximation.
%Additionally, even though the optimal algorithm for $L=1, K=1$ above is a (pseudo) polynomial time algorithm, it is far too slow to make metareasoning decisions guiding a heuristic search algorithm.

Let $P_0^\prime$ be
the success probability of instance ${\cal I}$  deciding not to dispatch any action, and under the assumption that computations do not provide information about any $M_i$.
Examining the algorithm for
$K=L=1$, we see that an action $a$ in $H_i$ is potentially dispatched only if it has a probability of success $P_i^\prime$ higher than $P_0^\prime$ (not dispatching 
an action).
The difference $P_i^\prime - P_0^\prime$ in probability of success is the gain for
dispatching $a$.
Alternately, one can do a computation $c_i$ first, and then decide on dispatching. This is worthwhile only if the expected
utility (measured in probability of success $P_i$) is increased on average vs. dispatching an action immediately by doing $c_i$ first.  The gain $P_i-max_j P_j^\prime$
is called the net value of information (VOI) for $c_i$ (at the initial state). In an optimal policy, an action $a$ is dispatched immediately only if
no $c_i$ has a positive net VOI.

A practical algorithm based on approximating the optimal policy would thus consider whether dispatching some action $a_i$ is beneficial (improves success probability), and given such an action, checks whether some computational action $c_i$ has a positive net VOI, in which case $a_i$ is not dispatched immediately. That is the gist of our proposed scheme.

Several complications hinder the implementation of this scheme. First,
the success probability values are computed exactly in \kdsae instances, but in fact the deadlines are also uncertain and
distributions are not independent,
making the success probability computed for a  \kdsae instance an incorrect estimate of the actual success probability, as well as too slow to use for metareasoning within a planner.
We also lack a good measurement model for the computations, and the perfect information assumption is also not realistic.
Overcoming these problems towards an approximate realization of this scheme
within a planner is described next.

\section{Implementation within the Planner}

The insight from the optimal solution of the restricted case is that we need to measure the 
expected gain for dispatching an action and to consider VOI of computation. Below we discuss how these are actually done in the planner.

\subsection{State-Space Modifications to Support Acting}

We assume that at any given point we have dispatched $m$ plan steps $\pi = [a_1..a_m]$  (where $m$ is initially 0), with $dispatch\_time[j]$ being the time at which step $j \in [1..m]$ is to be executed. All states on the open list begin with these $m$ steps, so the resulting planning task is equivalent to search from the state reached by those $m$ steps, subject to states respecting temporal side-constraints as described next.

First, in the situated temporal planners, the STP used to capture the temporal constraints on a plan \cite{DBLP:conf/aips/CashmoreCCKMR18}, included the temporal constraints required in OPTIC and also required $t(a_j) \geq t_{now}$ for all plan steps --- because execution cannot start earlier than $t_{now}$ (the current time). In our case, as the first $m$ actions have been dispatched --- so can go `before now' --- we keep the temporal constraints required in OPTIC but instead also require:
$$\begin{array}{rll}
\forall{a_j \in H_i} & t(a_j) = dispatch\_time[j] & \mbox{if}\; j \leq m\\
& t(a_j) \geq {t_\mathit{now}} & \mbox{otherwise}
\end{array}$$

Second, an STP is additionally used to find the $dispatch\_time$ values themselves. If at time $t_{now}$ the decision is made that the $m+1{\mbox{th}}$ plan step to be dispatched is the snap action $a_{m+1}$, we take the STP that would be built for the state reached by the snap action sequence $[a_1..a_{m+1}]$ and set $dispatch\_time[m+1]$ to the earliest feasible value $\mathit{tmin}(a_{m+1})$ of step $a_{m+1}$ in this STP: the earliest time it can occur considering the ordering constraints between plan steps and the dispatch times of the previous steps.

Third, we redefine the notion of the `latest start time' for states. The scalar value described in Section~\ref{sec:metareasoninginanplanner} is defined with respect to all plan steps in an STP. As this would now include the $m$ steps that have been dispatched, we are instead interested in a latest start time that is conditional on $m$, i.e. what is the latest time we must dispatch step $m+1$:
$$\mathit{latest\_start}(H_i,m) = \mathop{\min}_{a_j \in H_i \mid j > m} \mathit{tmax}(a_j) $$

Finally, we must consider the consequences of action dispatch on the open list, and on duplicate state detection (which in OPTIC and prior work is through maintaining a set of memoized states). The open list issue is easy: if we dispatch the snap action $a_m$ as step $m$, then we remove from the open list every state whose plan prefix $H_i$ does not have $a_m$ as step $m$. For duplicate detection, we must be more careful: \fix{remove bullets to save space?}
\begin{itemize}
    \item We identify a set of states to `un-memoize': any memoized state whose plan prefix $H_i$ does not begin with $[a_1..a_{m+1}]$, and remove them from the memoized states.
    \item We add to the open list for re-expansion any state $S_i$ expanded earlier in search, whose plan prefix $H_i$ does begin with $[a_1..a_{m+1}]$, and for which one or more of its successor states was pruned due to being equivalent to one of these un-memoized states.
\end{itemize}

\subsection{Dispatch Estimates during Search}

Having discussed modifications to the state space, we now turn to how to make metareasoning decisions in the planner. If we have dispatched the plan steps $\pi = [a_1..a_m]$, then our metareasoning decision is either to not dispatch something yet or to dispatch {\em now} one of the snap-actions $\mathit{next} = [\alpha_1..\alpha_{n}]$ applicable in the state reached by $\pi$. We need to assess the utility of each of these possibilities, i.e. the probability of success in each case, which we assume to be related to its LPF via $1 - e^{LPF}$. We denote the probability of success for not dispatching, and for each of $\mathit{next}$, as $P_{Nd}$ and $P_{d1}..P_{dn}$, respectively.

We approximate these utilities by simulating for each case what would be the allocations of the computation time to processes in the ensuing search, over a simulated open list $\mathit{sim\_open}$. When estimating $P_{Nd}$, we use $\mathit{open}$: the open list at the current moment in search. For $P_k \in [P_{d1}..P_{dn}]$, we consider only the nodes on the open list whose plan step $m+1$ is $\alpha_k$:
\[
\mathop{sim\_open}(\mathit{open},\alpha_k)
= [H_i \in \mathit{open} \mid H_i[m+1] = \alpha_k  ]
\]

We then compute context-dependent LPFs for nodes on the simulated open list, with a context $c$ being the number of dispatched steps: $c = m$ for $P_{Nd}$, $c = m+1$ otherwise. This is to use the appropriate $\mathit{latest\_start}$ estimates: when calculating $\mathit{LPF}^c_i$ we use the same calculation as Eq \ref{eq:LPFs}, except we base $D_i$ on $\mathit{latest\_start}(H_i,c)$. Thus, $c=m+1$ means the dispatch options benefit from step $m+1$ having been dispatched. We approximate the LPF of allocating time to processes $p_1 ... p_n$  in this order, starting at time $t_b$, with $t_b = t_{\mathit{now}}$ for each of $P_{d1}...P_{dn}$ (we would dispatch {\em now}); and $t_b = t_{\mathit{now}} + t_{\mathit{wait}}$ for $P_{Nd}$, as `not dispatching' means waiting some amount of time. We assume $t_{\mathit{wait}} = t_u$. Then, the LPF under context $c$ of an open list is:
%
% $LPF([p_1, \ldots, p_n], t_b) = LPF_1(\mathbb{E}(M_1), t_b) + LPF( [p_2, \ldots, p_n], t_b + \mathbb{E}(M_1))$ where $LPF([p_1], t_b) = LPF_1(\infty, t_b)$.
%
% \begin{align*}
% &\begin{cases}
% LPF_1(\mathbb{E}(M_1), t_b) + \\
% \hspace*{20pt} LPF( [p_2, \ldots, p_n], t_b + \mathbb{E}(M_1))  & i > 1\\
% LPF_1(\infty, t_b) & i = 1\\
% \end{cases}
% \end{align*}
%
%
$$\mathit{LPF}^c(p_1..p_n) = \mathop{\sum}_{i \in [1..n]} \mathit{LPF}^c_i\big(\mathbb{E}(M_i), t_b + \mathop{\sum}_{j \in [1..i]} \mathbb{E}(M_j)\big)$$

This is an imprecise (but empirically informative) measure in a number of important regards. First, we assume each node $p_i$ is allocated $e_i$ expansions -- the expected number of expansions for it to reach the goal. This is reasonable if $p_1$ is the best option for reaching the goals, $p_2$ is the second-best option, and so on; which in reality, would require {\em inter alia} a perfect heuristic. Second, the order in which processes are considered is crucial: we fix the order here according to a snapshot of $\mathit{open}$, whereas in search proper, the order is due to the $Q$ values of processes, which in turn are a function of the time at which computation is to be allocated to them.

\subsection{Metareasoning for Action Dispatch Decisions}

%\subsubsection{Andrew's dispatch rules}

Having defined how we estimate LPFs for candidate options (not dispatching yet, or dispatching some action) we now formulate dispatch rules. A na{\"i}ve rule would be `dispatch the best $\alpha_k$ where $P_{dk} > P_{Nd}$'. Since we only have 
inherently noisy and possibly biased estimates (due to expansion delay, a global one-step-error estimate, and an imperfect heuristic), rather than actual probabilities,  we must be wary of dispatching an action if the benefit of doing so is small, and/or little search has been performed in the subtree reached by $\alpha_k$ in order to substantiate its probability approximation.

To address these issues, we first identify dispatch candidates $\mathit{explored} \subseteq \mathit{next}$, where $\alpha_k \in \mathit{explored}$ if a minimum number of expansions have been performed in its subtree, and $\mathit{sim\_open}(\mathit{open},\alpha_k)$ is of a minimum size (these are hyperparameters). Then, if $\mathit{explored} \neq \emptyset$ we find:
$$\alpha_{x} = \argmin_{\alpha_k \in \mathit{explored}} P_{dk}$$
and dispatch $\alpha_{x}$ if $P_{dx} > P_{Nd} + \mathit{dt}$ where $dt$ is our {\em dispatch threshold}.
A caveat of this, however, is that if search is predominantly focusing on the subtree beneath only a subset of $\mathit{next}$ (which is typically the case for heuristic search), any $\alpha_k \not\in \mathit{explored}$ will never be considered for expansion, even if their probabilities of success are ostensibly better. To address this, we additionally find:
$$\alpha_{y} = \argmin_{\alpha_k \in (\mathit{next} \setminus \mathit{unexplored})} P_{dk}$$
and if $P_{dy} > P_{Nd} + \mathit{sft}$, where $\mathit{sft}$ is our {\em subtree focus threshold}, we constrain search to only expand nodes beneath $\alpha_y$. This embodies common sense, insofar as while we do not want to dispatch an action with nebulous evidence, there is an intuitive value of information argument.

% The goal here is to provide a theoretical foundation
% for these intuitions, and help select
% thresholds justified by decision-theory.
% Note that information-obtaining conditional policies like the 3rd policy above are outside scope~of the \cope~model, so we need to suggest
% another scheme.

\begin{figure}[tb]
    \centering
    \includegraphics[width=0.9\linewidth]{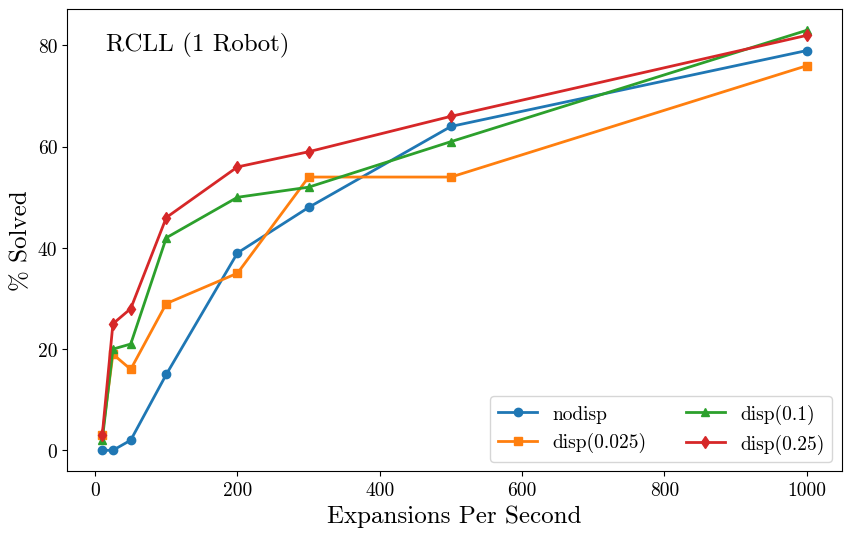}
    \includegraphics[width=0.9\linewidth]{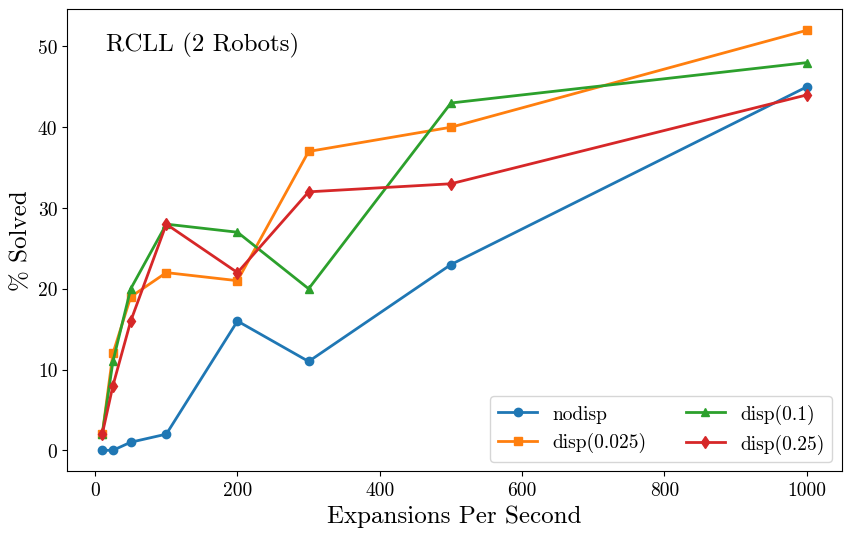}
    \includegraphics[width=0.9\linewidth]{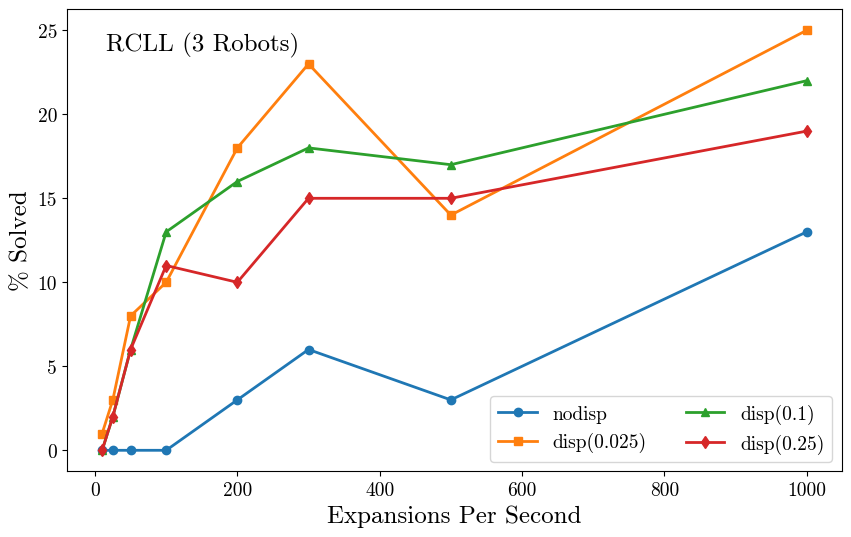}
    % \includegraphics[width=1\linewidth]{Turtlebot.png}
    % (a) Turtlebot
    \caption{Problems solved vs. expansions-per-sec: RCLL}
    \label{f:plots}
    \vspace{-4mm}
\end{figure}

\section{Empirical Evaluation}

Concurrent planning and execution is meant to be used in situations with time pressure. %However, we can not prove that concurrent planning and execution is always better than situated temporal planning, and so we conducted an empirical evaluation to examine this claim. 
Unfortunately, most  International Planning Competition domains were designed for offline planning and do not have inherent time pressure. Therefore, in our evaluation we focus on domains inspired by real problems involving robots:  planning problems from the Robocup Logistics League (RCLL) planning competition \cite{RCLL-Planning} with one, two, and three robots, as well as planning problems for a Turtlebot performing an office delivery task. These were both used to evaluate situated temporal planning \cite{DBLP:conf/aips/ShperbergCKRS21}.
The results for Turtlebot are not interesting -- nearly every instance is solved by every configuration, and thus these results are relegated to the supplementary material.
% However, since every tested algorithm successfully solved all Turtlebot instances across every configuration, we have chosen to exclude these results from the paper.
%We also modify IPC problems to feature extreme time pressure by assigning a deadline to when the first action must be dispatched.

\fix{how was time pressure ensured?  we claim that these instances have strong time pressure - what is evidence?}

We compare our concurrent planning and execution approach, denoted by {\em disp}, to the situated temporal planner \cite{DBLP:conf/aips/ShperbergCKRS21} using its default parameters, denoted by {\em nodisp}. As both approaches use the same situated temporal planner, both are equally informed, thus allowing for a clean evaluation of the value of allowing dispatching actions before search completes, as well as the ability of our metareasoning approach to make these decisions correctly. Most parameters for {\em disp} are the same default parameters used for {\em nodisp}. For the new parameters introduced for making dispatching decisions, we set the minimum number of expanded nodes in the subtree for dispatching and the minimum number of nodes in $sim\_open$ for expansion to 10. We vary the dispatch threshold (trying 0.25, 0.1, and 0.025), and set the subtree focus threshold to half the dispatch threshold. Ablation studies with other parameter settings (described in the supplementary material) show that these settings do capture a VOI criterion, which is important to the performance of our planner, as suggested by our theoretical results.

% evaluate the following configurations:
% \begin{tabular}{|c|c|c|}
%      dispatch threhold &  subtree focus threshold & min expansions\\
%      \hline
%      0.25 & 0.125 & 10\\
%      0.1 & 0.05 & 10\\
%      0.025 & 0.0125 & 10\\
% \end{tabular}
% }

Furthermore, because both approaches rely on the same planner, we choose to measure planning time by the number of expanded nodes rather than real wall-clock time. This allows for reducing random noise due to timing issues and makes all of our experiments deterministic. Specifically, we use a user-specified parameter specifying how many node expansions we can perform per second and then timing in the planner is based on the number of nodes expanded so far (divided by this parameter) rather than on the time that actually elapsed. We can then vary the number of expansions per second, simulating different `CPU speeds'.

Figure \ref{f:plots} shows the number of problems solved by both approaches for different CPU speeds. As the results show, when the CPU speed is low (meaning fewer nodes can be expanded before the deadline expires), the benefit from using concurrent planning and execution is high. When the CPU is fast enough, the difference decreases. Furthermore, in cases where the CPU exhibits sufficient speed to solve the problem before the deadline, it is plausible that the no-dispatch strategy could surpass all dispatching policies, which may commit to a suboptimal action. 
% This phenomenon becomes apparent, for instance, in the case of two robots when the rate of expansions per second exceeds 380.
This is in line with the theoretical limit of an infinitely fast CPU, where the best approach is to use offline planning to find a complete plan, then start executing it at time 0. On the other hand, CPU speeds of robotic space explorers are typically at least an order of magnitude less than CPU speeds of computers on Earth, and thus we believe this is one area where our approach can be useful.

\section{Discussion and Future Work}

We have formalized the \modelname ~metareasoning problem, aimed at bridging the gap between planning and acting concurrently.
%This can serve as a starting point for follow-up research on this topic. 
Insights from
optimal solutions of a restricted version of \modelname ~are used in a new metareasoning algorithm deciding whether to dispatch an action during search. 
Despite having only rudimentary probability estimates and no realistic measurement model (improving them is an important issue for future work),
our empirical evaluation shows that our algorithm works well under time pressure. In future work, we intend to develop a way to identify whether a current problem instance has strong time pressure, thus automatically switching between concurrent planning and execution, situated planning, and offline planning. More generally, we will explore the connection to other types of thinking fast and slow in AI \cite{DBLP:conf/aaai/BoochFHKLLLMMRS21}.

We are still a long way from the integration of our algorithm with real robots. First, there are technical challenges involved with actually dispatching an action on real hardware.
Second, the real world features uncertainty. Actions can fail, or take longer or shorter than expected to execute, and the executive must handle these. Even introducing actions with controllable durations raises questions about dispatchability \cite{DBLP:conf/aips/Morris16} that need to be addressed.
One possibility is to leverage the replanning compilation suggested for situated temporal planning \cite{DBLP:conf/aips/CashmoreCCKMR19}, which can also be used for our concurrent planning and execution formulation. Additionally, uncertain action durations can be handled as part of the planning process \cite{DBLP:journals/ai/CimattiDMRS18}. Thus, we believe the framework presented here can serve as a principled basis for an executive that can be used on real robots.

%\fix{real execution, uncertainty, action failure, replanning, ...}

\section*{Acknowledgements}

This research was supported by Grant No.~2019730 from the United States-Israel Binational Science Foundation (BSF) and by Grant No.~2008594 from the United States National Science Foundation (NSF).
The project was also funded by the EPSRC-funded project COHERENT (EP/V062506/1), the Israeli CHE Data Science Grant, by a Grant from the GIF, the German-Israeli Foundation for Scientific Research and Development, by ISF grant \#909/23, by MOST grant \#1001706842,
and by the Frankel center for CS at BGU.

\bibliography{main}

\end{document}

\maketitle

This document contains pseudo-code and ablation studies
% and full experimental results 
that were not able to be included in the main paper due to space constraints.

\section{Pseudo-code}

This appendix includes the pseudo-code for solving the special cases of \modelname, described in Section 4. Algorithm~\ref{alg:speical-case} solves the case of $K=0$ and $L=1$, whereas Algoirithm~\ref{alg:full} solves the case of $K=L=1$. 

\begin{algorithm}[tbh]
\caption{Solving $\mathcal{I}_i$ ($K=0$, $L=1$)}
\label{alg:speical-case}
\begin{algorithmic}[1]
        \STATE {\bf function} Solve01($\mathcal{I}_i$)
	\STATE Max$P \leftarrow 0$
    \FOR{$j=1$ to $n$}
        \STATE $P_j \leftarrow 0$
        \FOR{$o \in \text{support}(m_j)$}
            \STATE $(P, \pi_{ij}) \leftarrow Solve00(\mathcal{I}^0_i(c_j, o)$) 
            \STATE $P_j \leftarrow P_j + P \cdot m_j(o)$
        \ENDFOR
        \IF{$\text{Max}P < P_j$}
        \STATE $\text{Max}P \leftarrow P_j$
        \STATE $\text{best\_}c_j \leftarrow c_j$
		\ENDIF
    \ENDFOR
    \RETURN $\text{Max}P, \text{best\_}c_j$
\end{algorithmic}
\end{algorithm}

\begin{algorithm}
\caption{Solving $\mathcal{I}$ ($K=1$, $L=1$)}
\label{alg:full}
\begin{algorithmic}[1]
	\STATE Max$P \leftarrow 0$
    \FOR[No dispatch at $t=0$]{$j=1$ to $n$}
        \STATE $P_j \leftarrow 0$
        \FOR{$o \in \text{support}(m_j)$}
            \STATE $(P, \pi_j) \leftarrow Solve00( \mathcal{I}(c_j, o))$ 
            \STATE $P_j \leftarrow P_j + P \cdot m_j(o) $
        \ENDFOR
		\IF{$\text{Max}P < P_j$}
        \STATE $\text{Max}P \leftarrow P_j$
        \STATE $\text{policy} \leftarrow [c_j]$
		\ENDIF
    \ENDFOR
     \FOR[Dispatch $a_i\in H_i$ at $t=0$]{$i=1$ to $n$}
     \STATE $(P_i^\prime, \text{best\_}c_j)\leftarrow Solve01(\mathcal{I}_i)$ 
    \IF{$\text{Max}P < P_i^\prime $}
    \STATE $\text{Max}P \leftarrow P_i^\prime $
    \STATE $\text{policy} \leftarrow [a_i,\text{best\_}c_j]$
    \ENDIF
    \ENDFOR
    \RETURN Max$P$,\text{policy}
\end{algorithmic}
\end{algorithm}

\section{Ablation Studies}

\begin{figure*}
    \centering

    \includegraphics[width=0.45\linewidth]{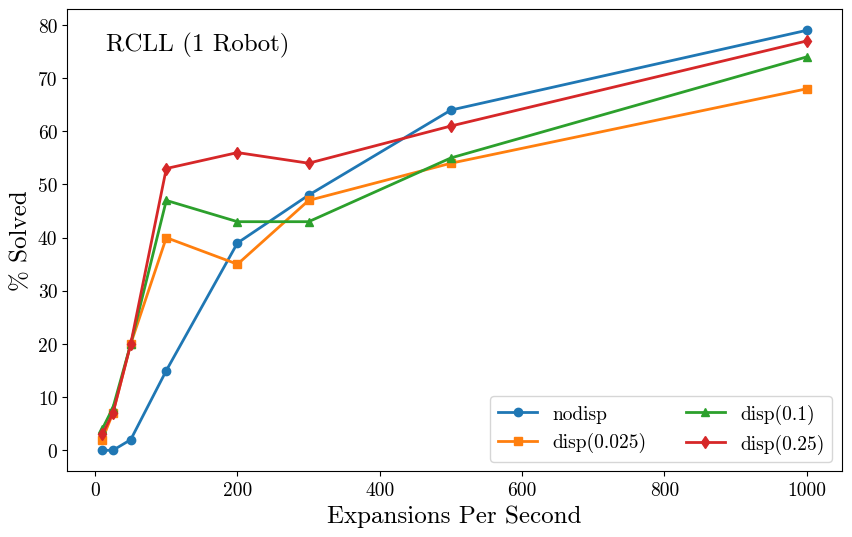}
    \includegraphics[width=0.45\linewidth]{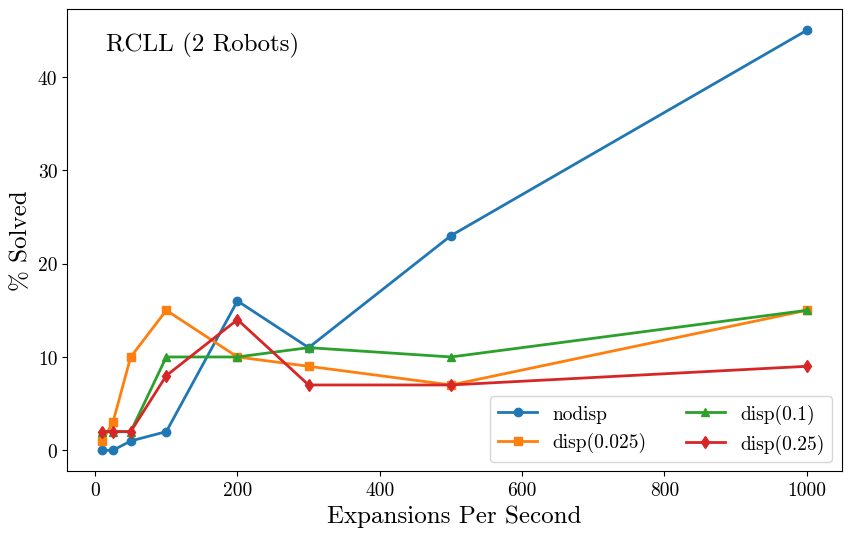}
    \includegraphics[width=0.45\linewidth]{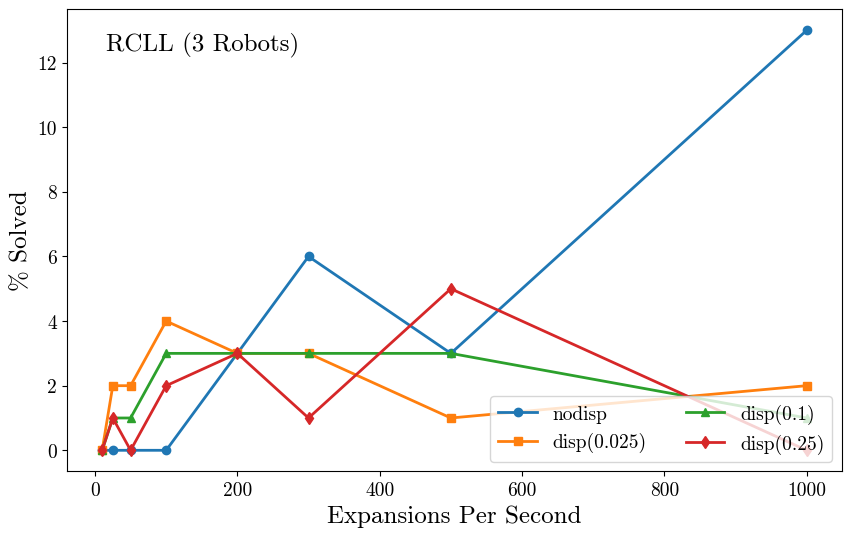}
    \includegraphics[width=0.45\linewidth]{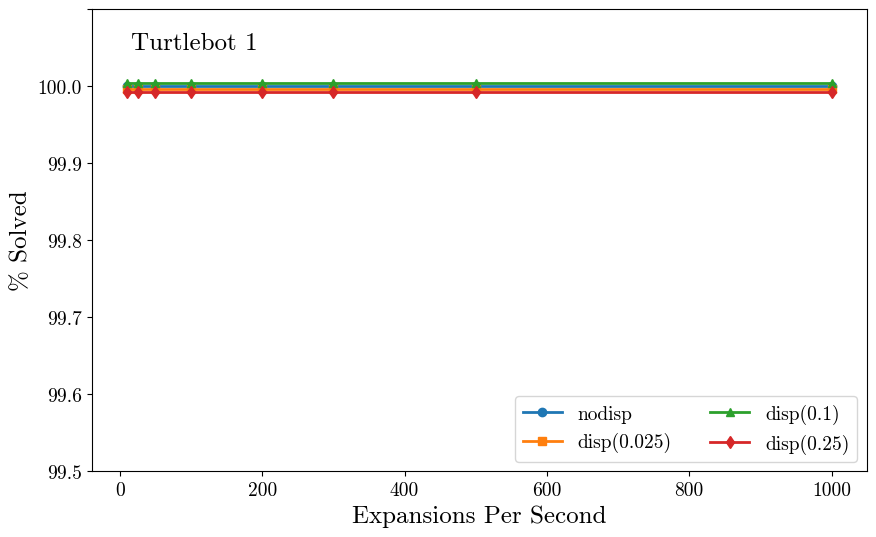}
    \includegraphics[width=0.45\linewidth]{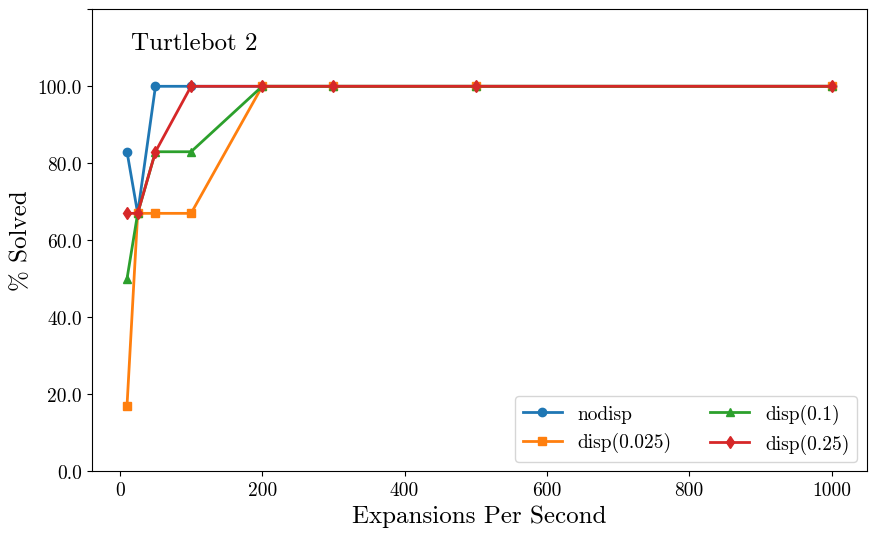}
    
    % Subtree focus threshold = 1, Dispatch frontier size	= 1
    % SFT - Subtree focus threshold 	
    % DFS - Dispatch frontier size	
    
    % DISP(10,0.25,0.125)
    % DISP(10,0.1,0.05)
    % DISP(10,0.025,0.0125)
    
    % DISP(1,0.25,1)
    % DISP(1,0.1,1)
    % DISP(1,0.025,1)

    % NODISP
    
    % \caption{Ablation 1: Minimum Expansions=1}
    \caption{Ablation 1: Dispatch frontier size = 1}
    \label{f:ablation1}
\end{figure*}

We performed three ablation studies. In the first one, we set the minimum number of expanded nodes in the subtree and in $sim\_open$ to 1, thereby allowing dispatching of actions based on uncertain estimates. The results for these are shown in Figure \ref{f:ablation1}. As the results show, {\em nodisp} performs better than {\em disp} with the different dispatch thresholds in RCLL for higher CPU speeds, showing that this VOI criterion is important.

\begin{figure*}
    \centering
    % \includegraphics{}
    
    % DISP(10,0.25,0.125)
    % DISP(10,0.1,0.05)
    % DISP(10,0.025,0.0125)

    \includegraphics[width=0.45\linewidth]{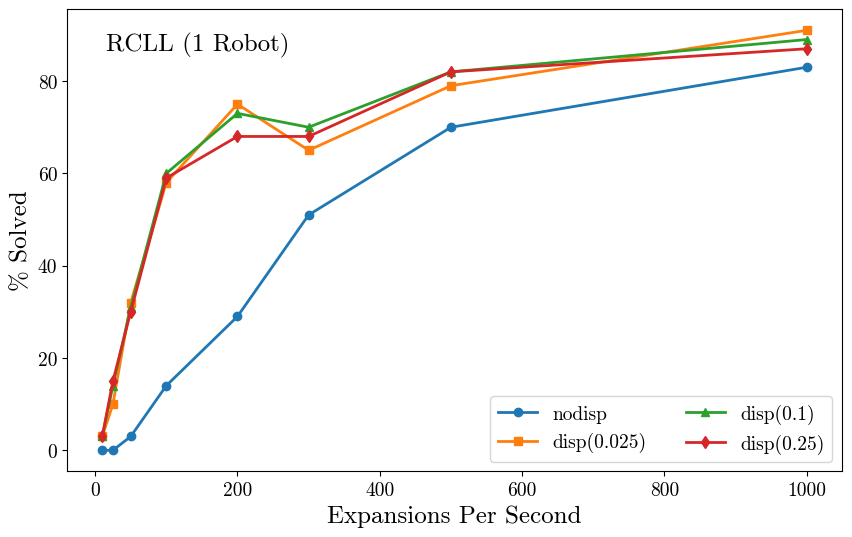}
    \includegraphics[width=0.45\linewidth]{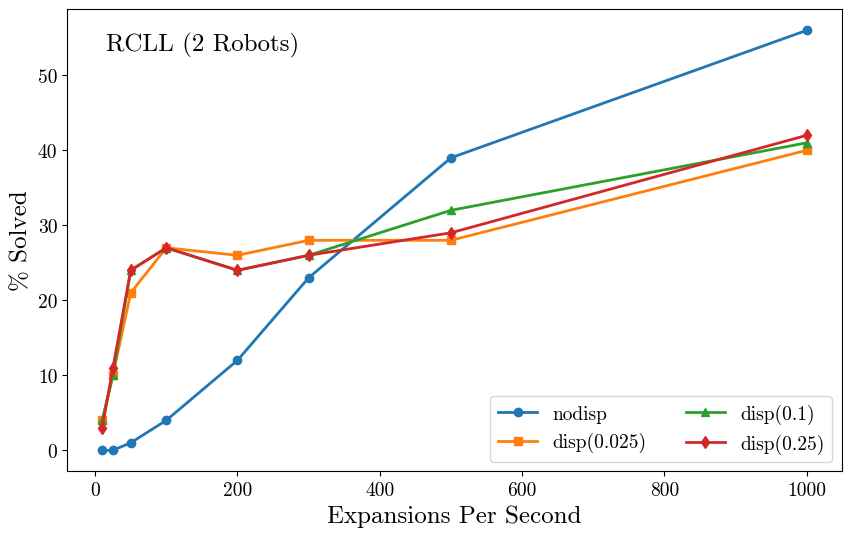}
    \includegraphics[width=0.45\linewidth]{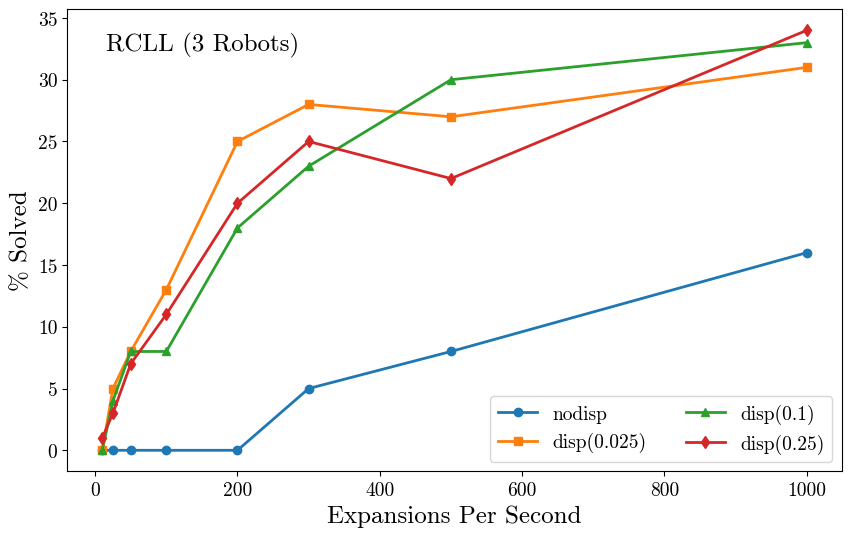}
    \includegraphics[width=0.45\linewidth]{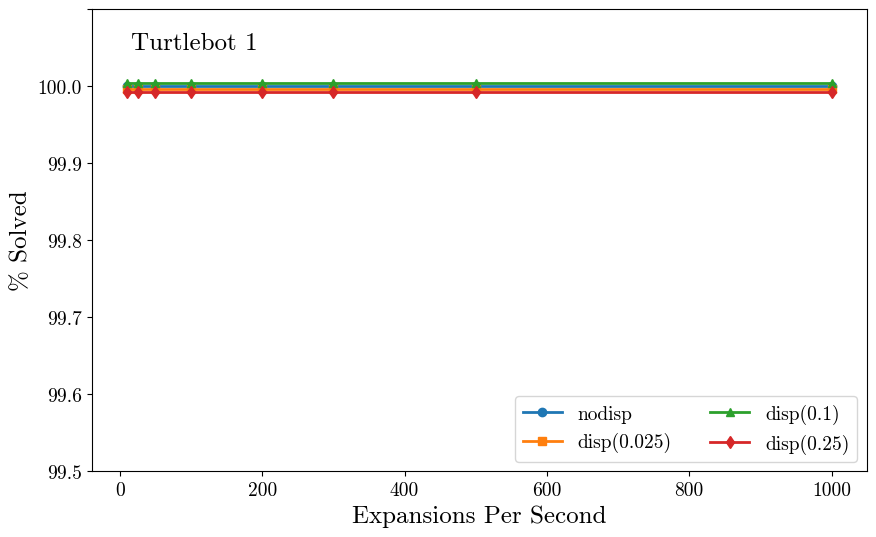}
    \includegraphics[width=0.45\linewidth]{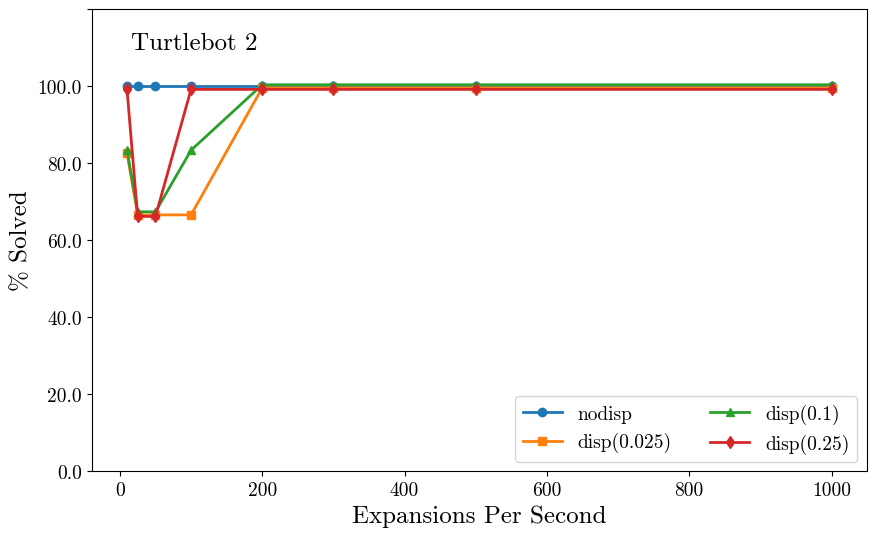}

    % SFT=10, DFS=1
    
    % DISP(1,0.25,1)
    % DISP(1,0.1,1)
    % DISP(1,0.025,1)

    % NODISP
    
    % \caption{Ablation 2: Minimum Expansions=1}
    \caption{Ablation 2: Subtree focus threshold = 1}
    \label{f:ablation2}
\end{figure*}

In the second ablation, we set the subtree focus threshold to 1, thereby eliminating the option to focus search on promising candidates which have not been explored enough. The results for these are in Figure \ref{f:ablation2}. While {\em disp} outperforms {\em nodisp} in most cases, in RCLL with 2 robots {\em nodisp} is better with higher CPU speeds. This is likely because our dispatching planner commits to a wrong decision, as the promising nodes have not been explored enough.

% \section{Full Experimental Results}

% Finally, we report the full results, including Turtlebot, in Figure \ref{f:full}. As these results show, 

% % In the third ablation, we set the subtree focus threshold to half of the dispatch threshold. The results for these are in Figure \ref{f:ablation3}.

% \begin{figure}
%     \centering
%     % \includegraphics{}

%     \includegraphics[width=0.45\linewidth]{figures/SFT_0.125_DFS_10/RCLL_1_SFT_0.125_DFS_10.png}
%     \includegraphics[width=0.45\linewidth]{figures/SFT_0.125_DFS_10/RCLL_2_SFT_0.125_DFS_10.png}
%     \includegraphics[width=0.45\linewidth]{figures/SFT_0.125_DFS_10/RCLL_3_SFT_0.125_DFS_10.png}
%     \includegraphics[width=0.45\linewidth]{figures/SFT_0.125_DFS_10/Turtlebot_1_SFT_0.125_DFS_10.png}
%     \includegraphics[width=0.45\linewidth]{figures/SFT_0.125_DFS_10/Turtlebot_2_SFT_0.125_DFS_10.png}
    
%     % Subtree Focus Threhold=1/2 Dispatch Threshold, Dispatch frontier size = 10
    
%     % DISP(10,0.25,0.125)
%     % DISP(10,0.25,1)
    
%     % DISP(10,0.1,0.05)
%     % DISP(10,0.1,1)
    
%     % DISP(10,0.025,0.0125)
%     % DISP(10,0.025,1)

%     % NODISP
%     \caption{Full Empirical Results}
%     \label{f:full}

% \end{figure}

% INCLUDE TABLES WITH RESULTS FOR TURTLEBOT, ALL PARAMETER SETTINGS

% Columns:
% Domain	
% EPS (Expansions per second)	
% Subtree focus threshold 	
% Dispatch frontier size	
% Dispatch threshold 	
% Total number of problems solved by disp	
% Total number of problems solved by nodisp	